# A Deep Learning Approach to Teeth Segmentation and Orientation from Panoramic X-Rays

**Mou Deb [1], Madhab Deb [2], and Mrinal Kanti Dhar [3,*]**

[1] Bioinformatics and Computational Biology, University of Minnesota-Twin Cities, Minneapolis, MN 55455, USA; deb00022@umn.edu
[2] Information and Communication Technology, Shahjalal University of Science and Technology, Sylhet 3114, Bangladesh; madhabgjb@gmail.com
[3] Department of Radiology, Mayo Clinic, Rochester, MN 55905, USA; dhar.mrinalkanti@mayo.edu
* Correspondence: dhar.mrinalkanti@mayo.edu

**Abstract**

Accurate teeth segmentation and orientation are fundamental in modern oral healthcare, enabling precise diagnosis, treatment planning, and dental implant design. In this study, we present a comprehensive approach to teeth segmentation and orientation from panoramic X-ray images, leveraging deep-learning techniques. We built an end-to-end instance segmentation network that uses an encoder–decoder architecture reinforced with grid-aware attention gates along the skip connections. We introduce oriented bounding box (OBB) generation through principal component analysis (PCA) for precise tooth orientation estimation. Evaluating our approach on the publicly available DNS dataset, comprising 543 panoramic X-ray images, we achieve the highest Intersection-over-Union (IoU) score of 82.43% and a Dice Similarity Coefficient (DSC) score of 90.37% among compared models in teeth instance segmentation. In OBB analysis, we obtain the Rotated IoU (RIoU) score of 82.82%. We also conduct detailed analyses of individual tooth labels and categorical performance, shedding light on strengths and weaknesses. The proposed model's accuracy and versatility offer promising prospects for improving dental diagnoses, treatment planning, and personalized healthcare in the oral domain. Our generated OBB coordinates and codes are available at https://github.com/mrinal054/Instance_teeth_segmentation.

## 1. Introduction

Accurate teeth segmentation is important in oral healthcare. It provides location data for orthodontic treatments, clinical diagnoses, and surgical procedures. It is also used to identify individuals, capture tooth morphology, and plan dental implants [1]. Manually segmenting teeth is time-consuming and challenging, even for experienced professionals. Semi-automatic techniques can help, but they still require some human input. This is especially true for lower-resolution images, with which it can be difficult to accurately delineate certain tooth regions.

Teeth segmentation from panoramic X-ray images can be of two types: semantic segmentation and instance segmentation. Semantic segmentation is a simple approach to dental analysis that labels the entire teeth region with a single label. This simplicity makes it computationally efficient and easy to implement. Semantic segmentation is well suited for general dental analysis for which distinguishing individual teeth is not essential, such as identifying overall dental health or disease patterns. Semantic segmentation could be used to track the progression of dental diseases over time. This would allow dentists to monitor the effectiveness of treatment and make adjustments as needed. Teeth instance segmentation assigns separate labels to each individual tooth, enabling precise identification of each tooth's boundaries and characteristics. With instance segmentation, dental professionals can accurately analyze the health of each tooth, aiding in personalized treatment planning for orthodontic procedures and other dental interventions. A dentist can use instance segmentation to identify a tooth that is decayed. This information can be used to plan the best course of treatment, such as a filling or a root canal. Instance segmentation allows for detailed quantitative analysis, such as measuring gaps between teeth, assessing wear patterns, and monitoring specific dental issues on a per-tooth basis. For tasks like designing dental prosthetics or implants, instance segmentation helps in creating accurate models that fit each tooth precisely.

While 3D instance segmentation is particularly valuable for complex cases that require a thorough understanding of tooth morphology, such as orthognathic surgery planning, dental implant placement, and in-depth orthodontic assessments, 2D instance segmentation is well suited for routine dental analyses and treatments that can be effectively assessed from a two-dimensional perspective, such as cavity detection, treatment planning, and basic orthodontic evaluations. Moreover, 2D instance segmentation is generally faster and requires less computational resources compared to its 3D counterpart, making it suitable for tasks that demand quick results.

In this paper, we use panoramic dental X-ray images for instance segmentation. Panoramic X-ray images are widely used for applications like dental caries, alveolar bone resorption, and impacted teeth [2]. Research works on panoramic X-ray images are mostly limited to teeth detection, teeth segmentation, and teeth numbering. Orientations of teeth are not well explored. It can help dentists and oral surgeons diagnose dental problems and plan treatments, particularly for procedures related to restorative dentistry, such as dental implants, teeth alignment, and orthodontic interventions. Typically, positioning technology and segmentation are conducted in separate frameworks. Consequently, the development cycle takes longer, and the algorithm's complexity is elevated [3].

Our contribution to this paper can be summarized as follows:

- We propose a two-stage framework that returns individual segmented teeth and their orientations. We adopt a deep learning-based approach to segment individual teeth, followed by applying Principal Component Analysis (PCA) to determine tooth orientation. Such outcomes can facilitate obtaining precise teeth positions in an image.
- We modify the FUSegNet model [4], originally developed for wound segmentation, for teeth instance segmentation by introducing grid-based attention gates in skip connections.
- We extend our approach to find horizontal bounding boxes (HBB), oriented bounding boxes (OBB), and missing teeth detection.
- We generate the OBB coordinates for each of the teeth in the DNS dataset [5], and these coordinates are available in our GitHub repository, as referenced in the *Data Availability Statement*.
- We provide comprehensive experimental findings, including a comparison with the state of the art, and an in-depth examination of our technique with a comprehensive ablation study to show the efficacy of our approach.

To highlight the motivation, approach, and impact of this study, we present a structured summary in Figure 1.

**Gap**

Studies typically focus on segmenting and numbering teeth, with limited exploration into the orientation of teeth.

**Solution**

- Deep learning-based instance segmentation
- PCA-based tooth orientation
- Public release of OBB annotations

**Clinical Implication**

- Pre-implant planning support
- Improved diagnosis
- Orthodontic assessment
- Missing teeth detection

**Figure 1.** Gap–solution–clinical implication framework. The figure highlights the lack of orientation analysis in existing studies, outlines our deep learning-based solution with PCA-guided orientation and OBB annotations, and summarizes the resulting clinical benefits.

## 2. Literature Review

Deep learning can help dentists by automating the process of teeth segmentation. This saves time and reduces human error, allowing dentists to focus on more important tasks such as accurate diagnosis and treatment planning. Various methods have been developed for teeth segmentation. Koch et al. [6] employed FCN in a U-Net setup. Zhao et al. [7] introduced TSASNet with a two-stage approach involving contextual attention and segmentation based on attention maps. Chen et al. [8] extended spatial pyramid pooling (SPP) to MSLPNet and introduced MS-SSIM loss. Salih and Kevin [9] proposed LTPEDN, replacing LBC layers with LTP layers. Hou et al. [10] proposed Teeth U-Net using a multi-scale feature aggregation attention block (MAB) and dilated hybrid self-attention block (DHAB) in the bottleneck layer for improved segmentation.

Jader et al. [11] are credited for being the pioneers who attempted teeth instance segmentation from panoramic X-ray images. They employed a set of 193 images for training and subsequently assessed their approach on 1224 images, achieving an F1-score of 88%. Rubiu et al. [12] also used Mask R-CNN on the Tufts Dental Database [13], which consists of 1000 panoramic dental radiographs, including both deciduous and permanent teeth. Their classification accuracy and dice score were 98.4% and 87%, respectively, and they observed poor segmentation for the right mandibular third molar. Silva et al. [5] explored Mask R-CNN, PANet, HTC, and ResNeSt for teeth segmentation and numbering. They observed the best result for PANet with 71.3% of mAP for segmentation. Helli and Mahamci [14] employed a two-step methodology in which they employed a U-Net to create a binary prediction, followed by morphological operations to label connecting elements. So, the full segmentation process is not deep learning-based. Their study utilized a limited dataset comprising 116 patients, with 11 images reserved for testing purposes. El Bsat et al. [15] tried MobileNet, AdapNet, DenseNet, and SegNet for maxillary teeth and palatal rugae segmentation. Their dataset consists of 797 occlusal views from teeth photographs. They achieved the best performance from SegNet with 86.66% average mIoU. Wathore and Gorthi introduced a bilateral symmetry-based augmentation technique for panoramic X-rays, achieving a Dice Similarity Coefficient (DSC) of 76.7% using TransUNet [16]. Similarly, Brahmi and Jdey presented a new dataset comprising 107 panoramic X-ray images, reporting an F1-score of 63% using Mask-RCNN [17].

The above-mentioned works primarily addressed teeth segmentation, with some including numbering, yet none tackled tooth orientation. However, the orientation of teeth in 2D panoramic X-ray images is significant because it can help dentists identify the position of the teeth and their roots. This paper presents a two-stage framework: initially, utilizing deep learning for precise teeth segmentation, enabling subsequent numbering based on individual labels; subsequently, employing principal component analysis (PCA) [18] to establish tooth orientation.

## 3. Materials and Methods

### 3.1. Dataset

In this paper, we use the publicly available DNS dataset, which is accessible upon request to the authors of [5]. The dataset comprises 543 panoramic X-ray images, split into five folds with a resolution of 1991 × 1127 pixels. The test fold has 111 images, while the other four have 108 each. Three folds are for training, and one is for validation. To fit in the GPU, we divide each image into patches of size 512 × 512 pixels with an overlap of 10 pixels along both height and width. This corresponds to a stride of 502 pixels, keeping redundancy low while enabling smooth transitions across patch boundaries. For border patches, the final starting index is adjusted by subtracting the patch size from the image dimension, ensuring that the patch fits entirely within the image boundaries. We choose the patch size 512 × 512 to balance contextual richness and computational efficiency. Larger patches preserve more semantic and structural information, which is beneficial for tasks such as teeth segmentation. All images are normalized between 0 and 1. In addition, for training the deep-learning models, a unique segmentation label is assigned to each tooth, resulting in a total of 32 segmentation labels. Table 1 summarizes the key properties of the DNS dataset and the preprocessing steps adopted for model training. These include image resolution, fold distribution, patching strategy, and the use of PCA-based OBB generation. The distribution of segmentation labels is depicted in Figure 2. Furthermore, we create oriented bounding box coordinates for each tooth using a PCA-based approach.

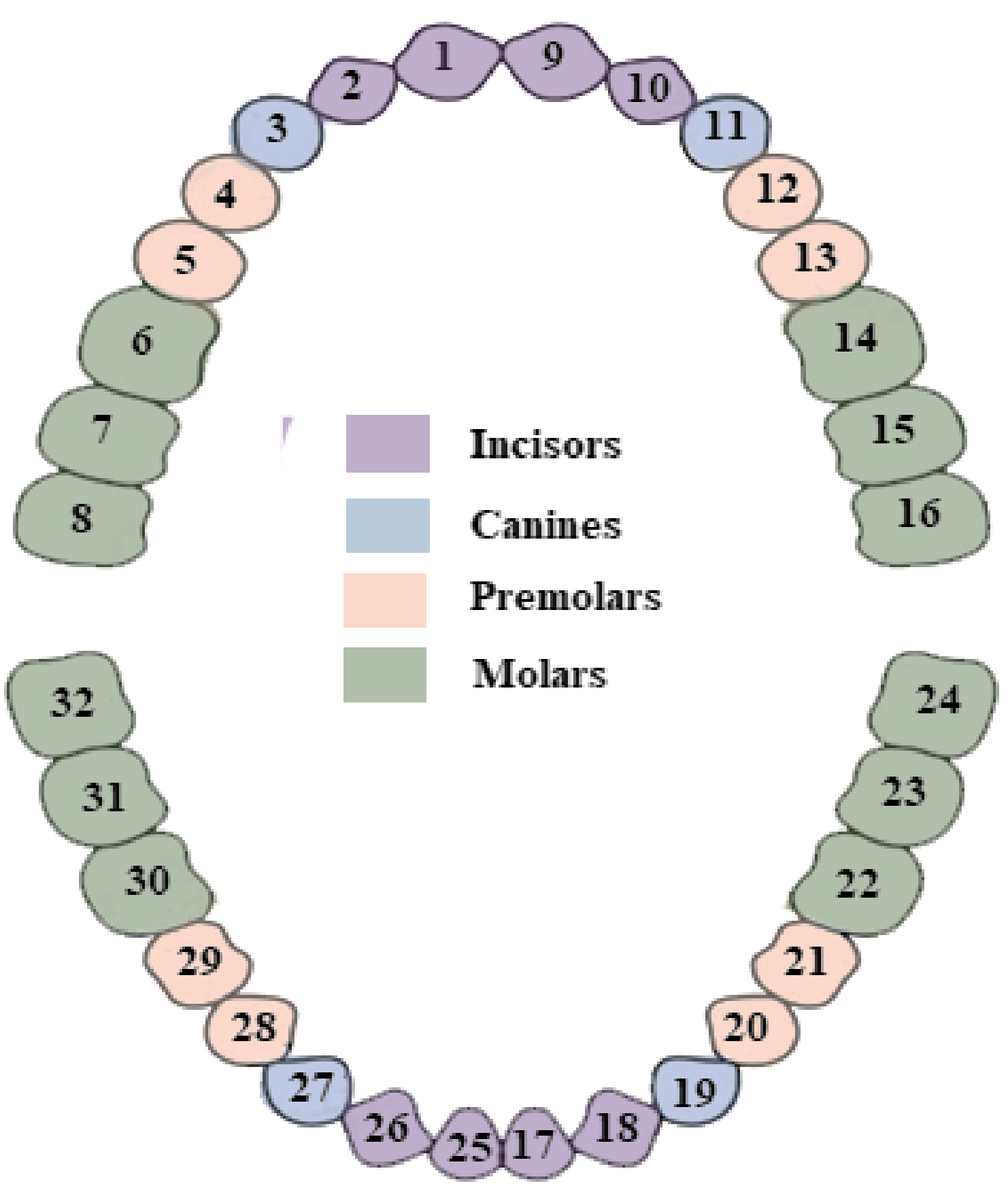

**Figure 2.** Tooth labeling used in this study. Each of the 32 teeth is assigned a unique label (1–32) and categorized into four anatomical groups: incisors, canines, premolars, and molars. Each group is further subdivided into upper and lower teeth, resulting in eight total categories. These categories are later used for categorical performance analysis (see Table 4 and Figure 11).

**Table 1.** Overview of the DNS Dataset and Preprocessing Steps.

| Attribute | Description |
|---|---|
| Dataset Name | DNS Dataset |
| Access | Publicly available upon request to the authors of [5] |
| Total Images | 543 panoramic X-ray images |
| Image Resolution | 1991 × 1127 pixels |
| Fold Distribution | 5 folds: 4 for training and validation, 1 for testing |
| Images per Fold | Test: 111 images; Other folds: 108 images each |
| Patch Size | 512 × 512 pixels |
| Patch Overlap | 10 × 10 pixels |
| Normalization | Pixel intensities scaled to [0, 1] |
| Segmentation Labels | 32 unique tooth labels |

### 3.2. Model Architecture

Figure 3 demonstrates our deep learning model. Unlike the original FUSegNet designed for wound segmentation, we introduce grid-based attention gates into the skip connections to enhance the model's focus on spatially relevant features, which is particularly beneficial for fine-grained dental structures. It has four major parts: encoder, decoder, grid-based attention gates (AGs), and parallel spatial and channel squeeze-and-excitation (P-scSE) module. Furthermore, to go beyond horizontal bounding boxes, which do not account for angular orientation, we implement a PCA-based oriented-bounding-box (OBB) technique, as detailed in Section 3.4.

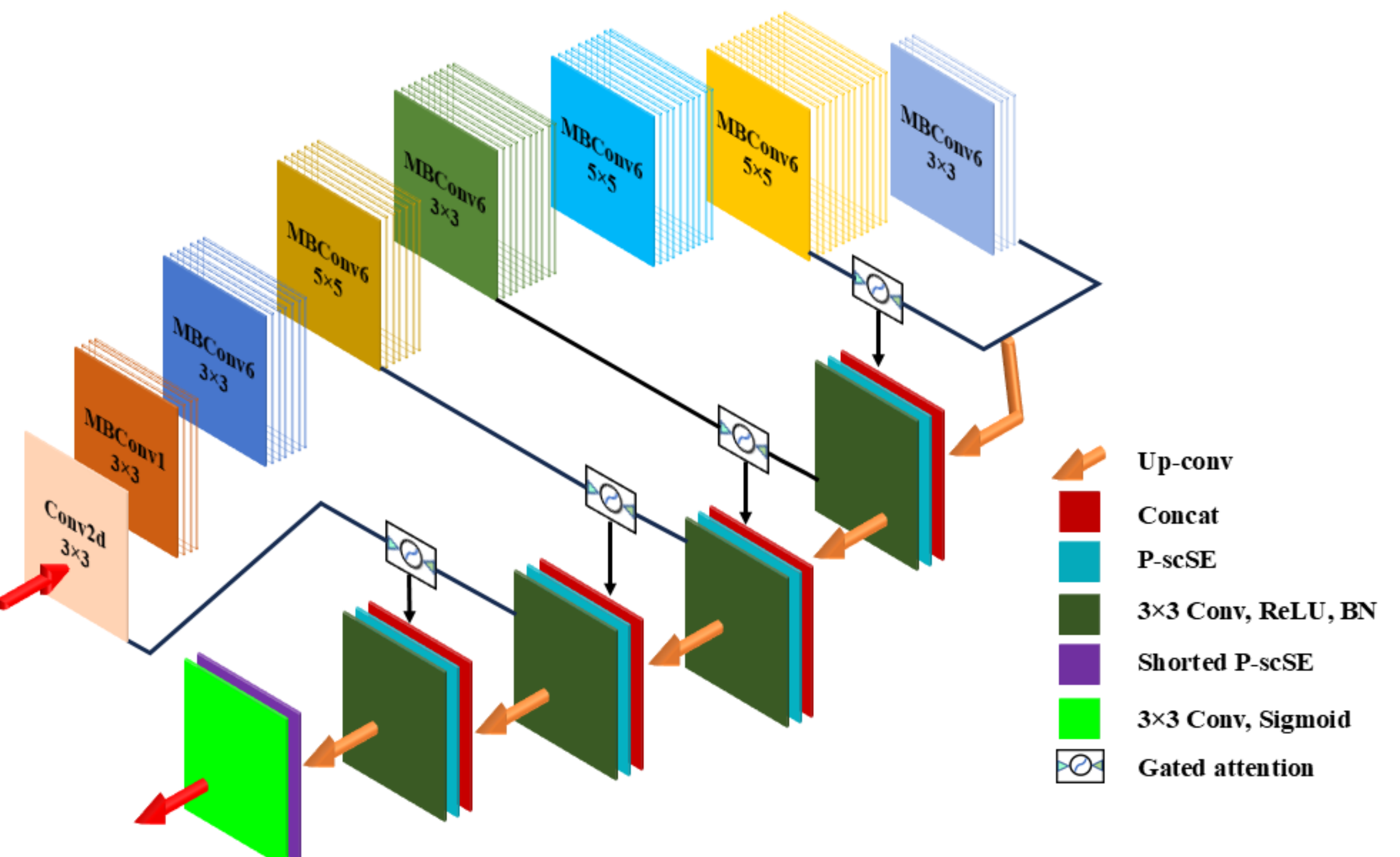

**Figure 3.** Overview of the proposed deep-learning model for teeth instance segmentation. It is a modified FUSegNet architecture with four key components: an EfficientNet-based encoder, a decoder with spatial and channel integration, grid-based attention gates for refined skip connections, and a P-scSE module for enhanced feature representation.

#### 3.2.1. Encoder

To avoid manual scale adjustments, the study employs an EfficientNet architecture as the core encoder. Convolutional neural networks often require tweaking depth, width, and resolution, a traditionally cumbersome and random process. EfficientNet's authors [19] introduce a novel approach, using fixed coefficients ($\alpha$, $\beta$, and $\gamma$) and a compound coefficient $\phi$ for uniform scaling. Depth, width, and resolution are scaled as $\alpha^{\phi}$, $\beta^{\phi}$, and $\gamma^{\phi}$, respectively. The authors used platform-aware neural architecture search via MnasNet [20] to discover the baseline EfficientNet-B0 architecture. Based on this, they conducted a small grid search to determine the compound scaling coefficients $\alpha$ = 1.2, β = 1.1, and γ = 1.15, which were used to uniformly scale the model to EfficientNet-B1 through B7. While EfficientNet-B0 is 224 × 224, EfficientNet-B7 has $\phi$ = 6, yielding 224 × $\gamma^{\phi}$=224 × $1.15^{6}$ ≈ 518 resolution. So, we use an EfficientNet-B7 model trained on ImageNet for patches of size 512 × 512.

#### 3.2.2. Decoder

As depicted in Figure 3, during each decoder stage, the upsampled output from the lower level is first concatenated with the encoder output of the corresponding level. However, in contrast to the FUSegNet architecture, the encoder output undergoes processing through a grid-based gated attention module [21]. This is performed to regulate the flow along the skip connections and allow the attention coefficients to focus more specifically on local regions. The resulting concatenated output then passes through the P-scSE attention module, which aggregates spatial and channel-wise information. Finally, a 3 × 3 Convolution-ReLU-Batch normalization is applied to this output.

#### 3.2.3. Parallel Spatial and Channel Squeeze-And-Excitation (P-scSE)

The squeeze-and-excitation module [22] was designed to boost the network's representational power by highlighting significant features and ignoring less relevant ones. It generates a channel descriptor using global average pooling, triggering channel-related dependencies. It is also referred to as cSE due to its channel-wise excitation. Roy et al. [23] introduced the sSE module, which squeezes along the channel axis and excites along spatial dimensions. The scSE module combines the cSE and sSE components. This combination can be conduted in different ways. As shown in Figure 4, P-scSE [4] creates two parallel branches of the scSE module: one by adding cSE and sSE and the other by taking the maximum of them. While max-out offers competitiveness between the channel and spatial excitations, addition aggregates these two excitations. A switch is provided to skip max-out when the number of channels is small. When we have a small number of channels, the model's capacity to learn intricate channel dependencies and patterns is already limited. So, if we selectively collect features, like max-out does, it will lose some important features and will not contribute significantly.

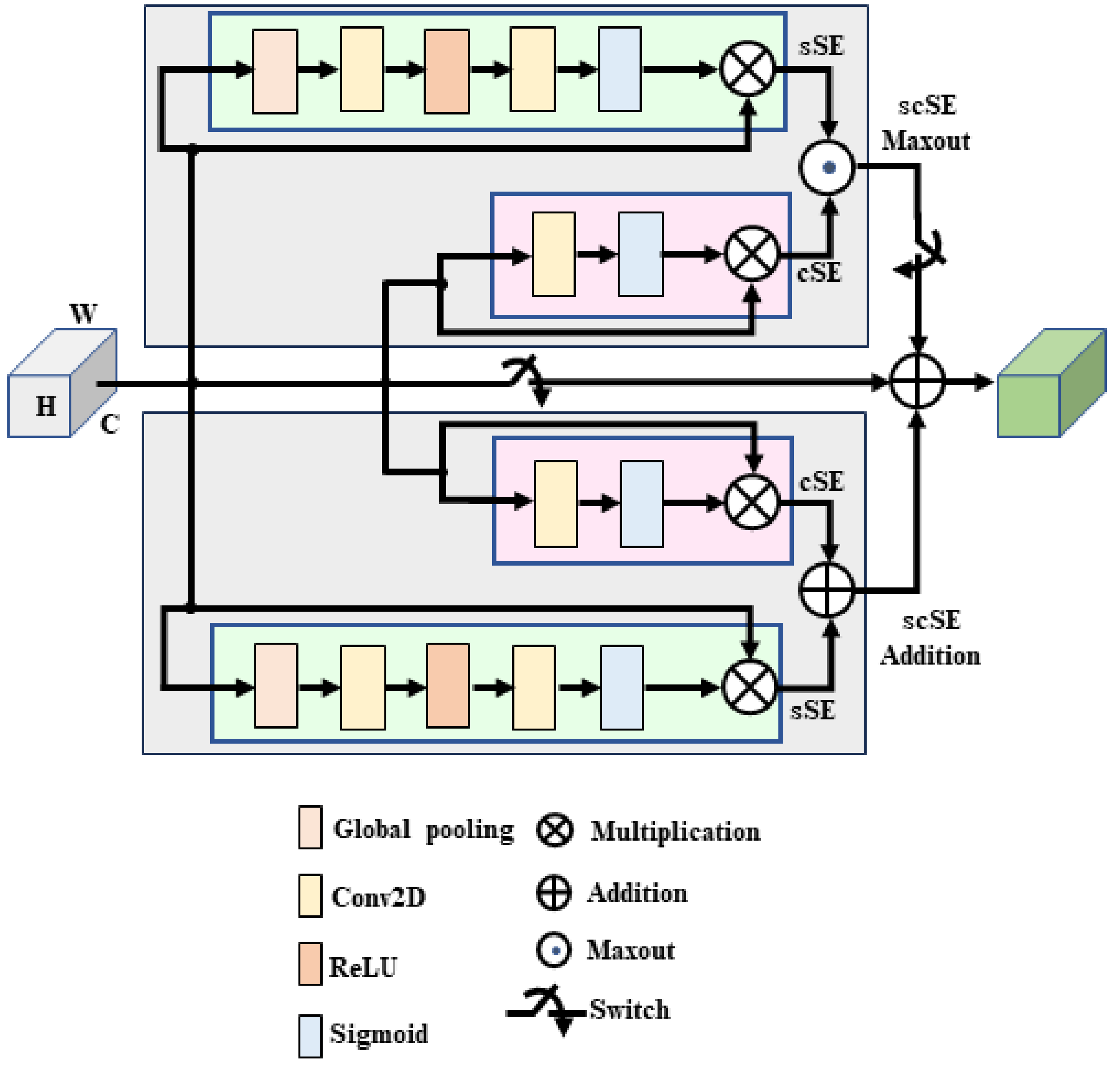

**Figure 4.** Architecture of parallel scSE (P-scSE).

#### 3.2.4. Grid-Based Attention Gate

In this paper, we use grid-based attention gates (AGs) [21] to improve skip connections. The AG is a trainable module that is added to skip connections of encoder–decoder-based architecture to reduce the number of false positive predictions for small objects with significant shape variation. It learns to weigh the features from the skip connections, giving more importance to the features that are relevant to the target structure. As shown in Figure 5, the attention gate calculates the attention coefficient $\alpha$, which is the result of additive attention computed from the input feature maps $x$ and the contextual information provided from the gating signal $g$. The attention coefficient identifies important spatial regions by paying attention to task-relevant activations.

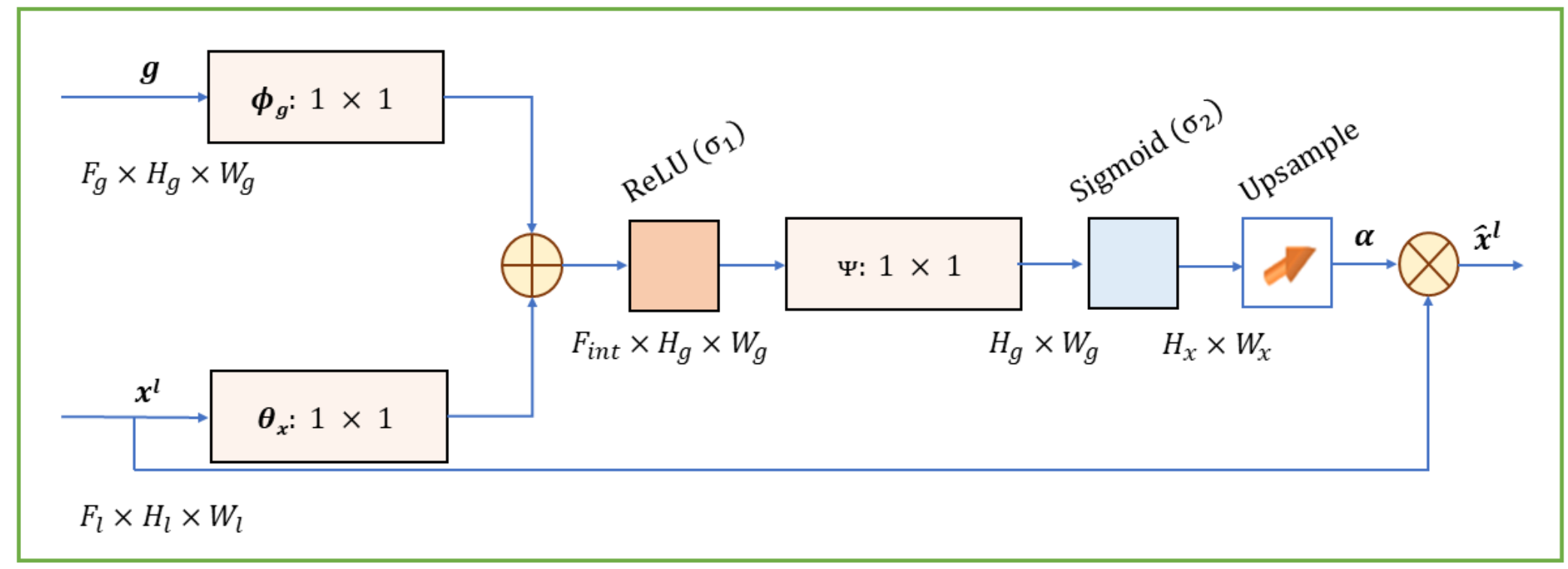

**Figure 5.** Architecture of the grid-based attention gate (AG). It is a 2D version of the AG shown in [21]. The attention coefficient $\alpha$ is computed using additive attention from the input feature maps (x) and gating signal (g), allowing the model to focus on spatially relevant features [24].

### *3.3. Post-Processing*

The output of the deep-learning model is fine-tuned by implementing a post-processing stage. If two regions with the same label exist, the larger region is identified as the desired region, while the smaller region is categorized as unwanted. Initially, we identify border pixels using chain coding. Subsequently, we detect neighboring pixels of each border pixel using 8-connectivity. We only consider those neighboring pixels with intensities differing from those of the unwanted portion. As shown in Figure 6, we observe three cases that need to be addressed.

*Case-I*: If all the neighbor pixels are 0 (background), then dissolve the unwanted portion into the background.

*Case-II*: If the neighbor pixels consist of background pixels and pixels of a specific label, then perform the following:

- First, ignore background pixels.
- Then, dissolve the unwanted portion into label pixels.

*Case-III*: If there exist two different labels in the neighboring pixels, then perform the following:

- Count the number of border pixels that lie in both labels.
- The most frequent label is the winner.

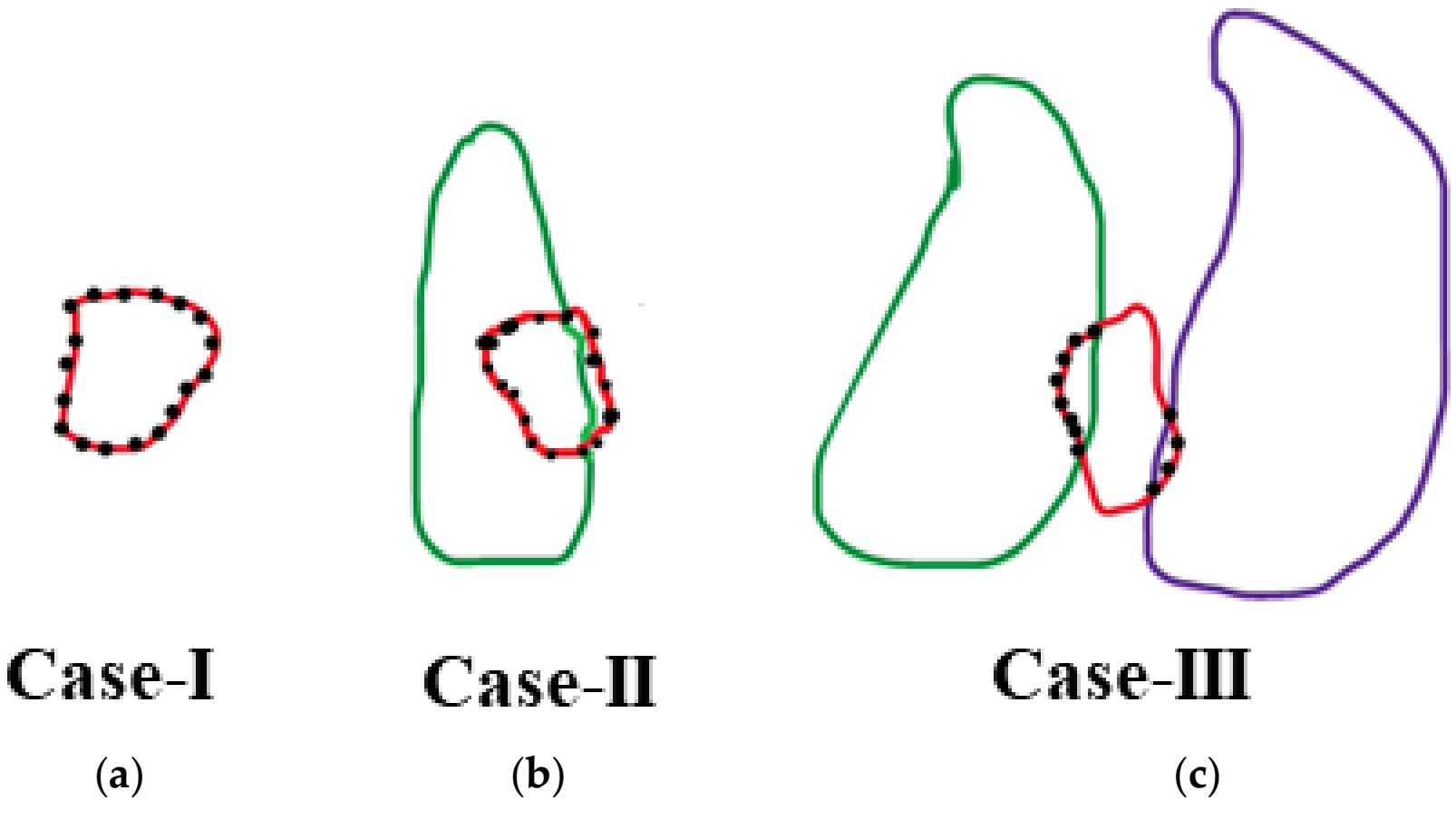

**Figure 6.** Post-processing cases to refine segmentation outputs. The red regions represent unwanted fragments with the same label as a nearby tooth. Black dots indicate border pixels of these unwanted regions. Three typical scenarios are illustrated: (**a**) Case-I: unwanted region surrounded entirely by background, which is removed; (**b**) Case-II: unwanted region adjacent to background and a labeled tooth, which is merged with the label; (**c**) Case-III: unwanted region bordering multiple labeled teeth, which is merged with the most frequent adjacent label.

### 3.4. Oriented Bounding Box (OBB)

To implement the oriented bounding box (OBB), we utilize principal component analysis (PCA) [18]. PCA is a statistical technique that transforms correlated observations into uncorrelated values called principal components. It ensures the first component captures the most variance, and subsequent components maximize variance while staying uncorrelated with preceding ones through orthogonal transformations.

We chose PCA due to its interpretability, computational efficiency, and effectiveness in capturing the primary axis of orientation from segmented tooth regions. While deep learning-based or regression-based OBB predictors are potential alternatives, they typically require supervised training with ground truth orientation angles, which are not available in the DNS dataset. We selected PCA over minimum-area bounding rectangles (e.g., OpenCV's minAreaRect) because, unlike minAreaRect, which returns a rotated rectangle without clear directional cues, PCA explicitly identifies the dominant direction of the tooth contour, making it more suitable for estimating rotation in our context. The steps involved in generating OBBs in this paper are as follows:

*Tooth separation*: We approach the OBB generation for each tooth individually. This involves keeping one tooth in the image while rendering the others as background. The image is then binarized.

*PCA*: We then calculate the first two major principal components (PCs) and determine the angle between the first PC and the horizontal axis, which we refer to as $PCA_{angle}$.

*Rotation*: We then generate a 2D rotation matrix to rotate the tooth and align it vertically. The rotation matrix for rotating a point ($x$, $y$) by an angle ($\theta$) around an arbitrary pivot ($x_c$, $y_c$) can be expressed as follows:

$$\begin{bmatrix} x' \\ y' \\ 1 \end{bmatrix} = \begin{bmatrix} \cos\theta & -\sin\theta & x_c(1-\cos\theta) + y_c\sin\theta \\ \sin\theta & \cos\theta & y_c(1-\cos\theta) - x_c\sin\theta \\ 0 & 0 & 1 \end{bmatrix} \begin{bmatrix} x \\ y \\ 1 \end{bmatrix} \tag{1}$$

We calculate $\theta$ in the following way:

$$\begin{cases} \theta = 180 + (90 - PCA_{angle}) & if\ PCA_{angle} < 0 \\ \theta = (90 - PCA_{angle}) & otherwise \end{cases} \tag{2}$$

*HBB*: After aligning the tooth vertically, we proceed to plot the horizontal bounding box (HBB). Generating an HBB from contour points is much easier than generating an OBB.

*Inverse rotation*: We then rotate the tooth along with the HBB back to its original position. To achieve this, we substitute $(-\theta)$ into Equation (1).

### 3.5. Training and Evaluation

Experimental setup: All experiments are executed on Google Colab Pro+. For optimization, we use the Adam optimizer [25] with an initial learning rate of 0.001 and weight decay of 1e-5. The learning rate is decreased by a factor of 0.1 when the metric shows no improvement for 10 epochs. Models are trained for 50 epochs, and the best model is stored. We employ a combined loss function, using both dice loss and focal loss with equal weights. Unlike cross-entropy loss, which overlooks contextual information of surrounding pixels in favor of per-pixel calculation, dice loss accounts for local and global information. Dice loss, represented as $DL = (1 - DSC)$, encompasses the dice coefficient ($DSC$). Focal loss ($FL$) is valuable for addressing class imbalance (such as background >> foreground), in which it prioritizes challenging cases by down-weighting easy examples [26]. So, the final loss function is expressed as follows:

$$\mathcal{L} = DL + FL \tag{3}$$

Evaluation metric: For the segmentation task, we use intersection-over-union (IoU), precision, recall, and dice score (DSC). For encoder–decoder-based architecture, these evaluation metrics are widely used. Here are each definition's details:

$$Precision = \frac{TP}{TP + FP} \quad (4)$$

$$Recall = \frac{TP}{TP + FN} \quad (5)$$

$$DSC = \frac{2TP}{2TP + FP + FN} \quad (6)$$

$$IoU = \frac{TP}{TP + FP + FN} \quad (7)$$

Here, *TP*, *FP*, and *FN* are true positive, false positive, and false negative, respectively.

For OBB, we calculate IoU between the ground truth OBB and predicted OBB. In later sections, it will be termed as rotated IoU (RIoU) to distinguish it from the IoU used in segmentation.

## 4. Results and Discussion

We can divide our experiments into two parts: segmentation analysis and OBB analysis. In segmentation analysis, we compare our proposed model with state-of-the-art segmentation models. Since our goal is to develop an encoder–decoder-based architecture, we primarily compare the results with popular encoder–decoder structures. We tabulate the results in Table 2. As shown in the table, our proposed model achieves an IoU score of 82.43% and a DSC score of 90.37%, both of which are the highest scores compared to the other models. Incorporating grid-based attention gates led to a ~1.5% IoU improvement over the original FUSegNet. The post-processing described in Section 3.3 improves the IoU from 82.0 to 82.43 and the DSC from 90.1 to 90.37. While the gain may appear modest, this step helps to remove artifacts and ensures cleaner segmentation masks, particularly around tooth boundaries.

We also explore two transformer-based approaches. The first one is the Mixed Vision Transformer (MiT)-b2-UNet, for which we incorporate the SegFormer (b2-sized) encoder [27] pretrained on ImageNet with the UNet architecture. The second one is Swin-Unet [28]. However, we do not observe satisfactory performance with transformer-based approaches. Possibly, it is because transformers often tend to overfit for small-scale data due to the lack of structural bias [29] or because they require additional task-specific tuning [30]. Figure 7 provides a graphical summary of the IoU, Precision, Recall, and DSC scores reported in Table 2, illustrating the overall performance of each model, with our proposed method achieving the highest scores across all metrics.

**Table 2.** Segmentation results achieved from deep-learning models.

| Model | IoU | Precision | Recall | DSC |
|---|---|---|---|---|
| DeepLabV3+ [31] | 80.45 | 88.59 | 89.75 | 89.17 |
| FPN [32] | 81.75 | 89.54 | 90.28 | 89.76 |
| MANet [33] | 62.70 | 85.82 | 69.95 | 77.08 |
| LinkNet [34] | 35.25 | 73.63 | 40.35 | 52.13 |
| FUSegNet [4] | 80.79 | 88.51 | 90.26 | 89.38 |
| PSPNet [35] | 28.94 | 59.60 | 36.00 | 44.89 |
| UNet [36] | 67.93 | 86.00 | 76.37 | 80.90 |
| MiT-b2-UNet [27] | 72.65 | 86.80 | 81.67 | 84.16 |
| Swin-Unet [28] | 53.69 | 72.43 | 67.47 | 69.87 |
| Ours | 82.43 | 90.48 | 90.26 | 90.37 |

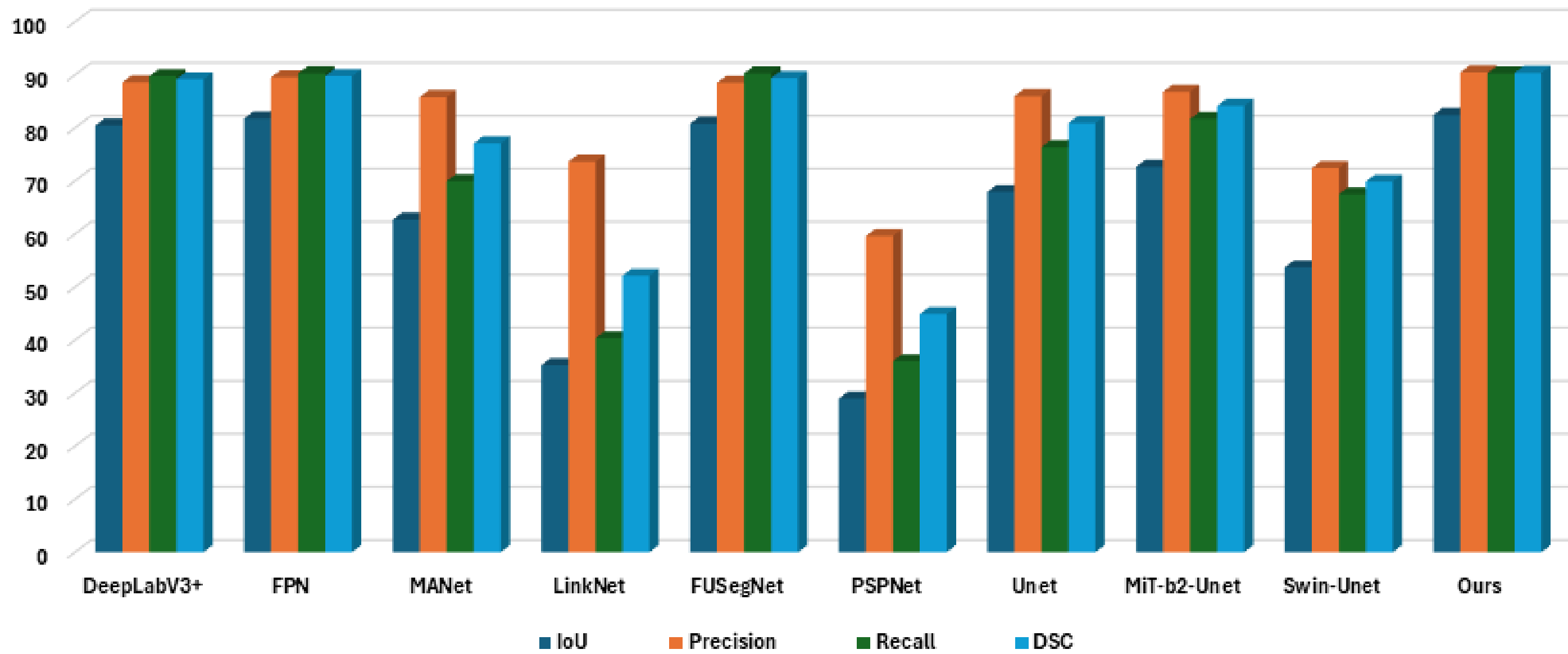

**Figure 7.** Bar chart comparison of segmentation performance across different deep-learning models. The proposed model achieves the highest scores in almost all evaluation metrics, indicating its superior performance.

Figure 8 illustrates the training and validation performance of the proposed model over 50 epochs. The model shows stable convergence with steadily decreasing loss and consistently high IoU and DSC scores on the validation set.

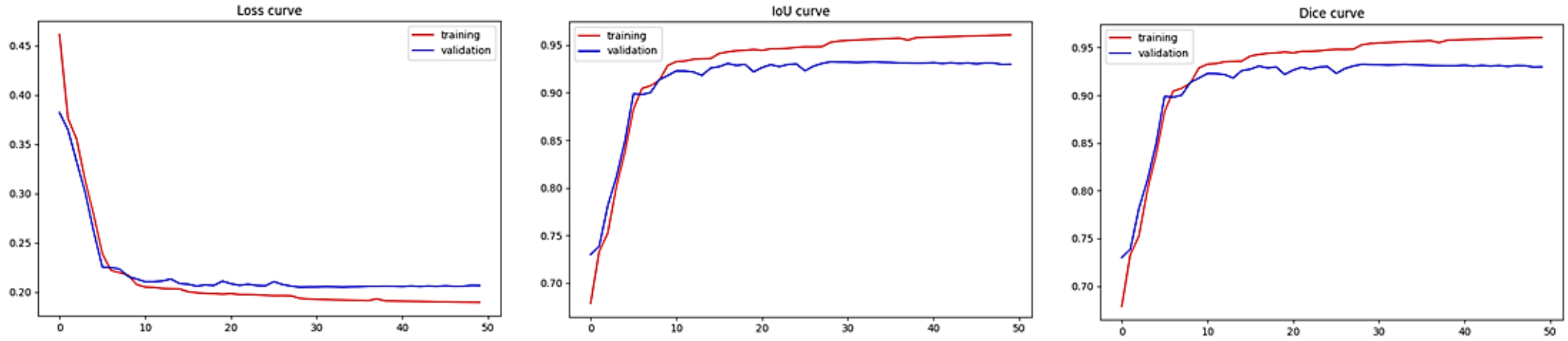

**Figure 8.** Training and validation performance of the proposed model over 50 epochs: (**left**) loss curve, (**middle**) Intersection-over-Union (IoU) curve, and (**right**) Dice Similarity Coefficient (DSC) curve.

For OBB performance analysis, we first generate oriented bounding boxes using the method described earlier and calculate rotated IoU (RIoU).

We prefer IoU over mean average precision because, unlike other region proposal-based detection methods, oriented bounding boxes generated from encoder–decoder-based approaches do not have any confidence scores. Figure 9 illustrates the $PCA_{angle}$ for several example teeth by overlaying the first principal component on each tooth shape. This angle, measured relative to the horizontal axis, is then used to compute the tooth rotation angle, $\theta$, for generating OBBs. As shown in Table 3, we achieve an RIoU score of 82.82%.

**Table 3.** Rotated IoU (RIoU), false positives (FP), and false negatives (FN) calculated from OBBs achieved with our proposed method.

| RIoU | FP | FN |
| --- | --- | --- |
| 82.82 | 0 | 3~6 |

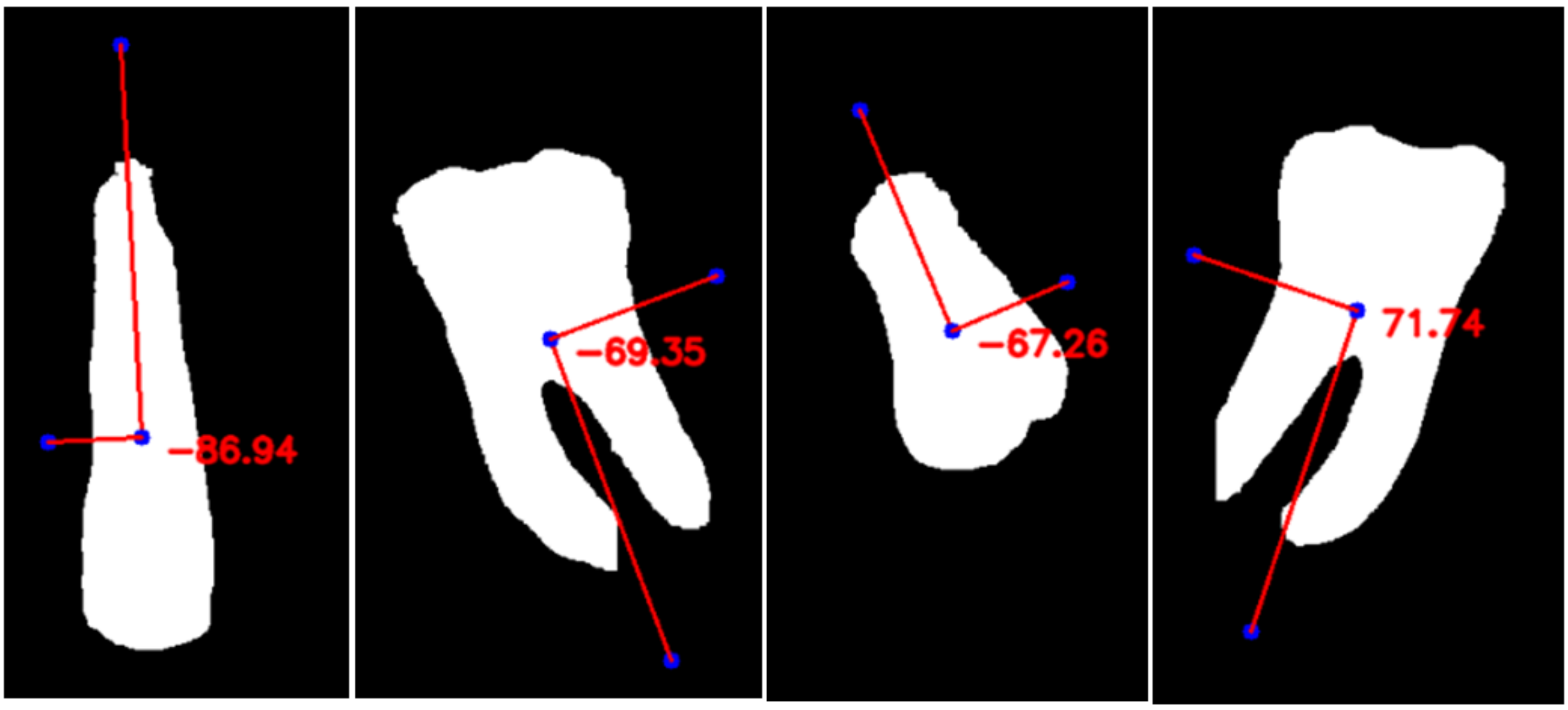

**Figure 9.** Visualization of $PCA_{Angle}$ for selected teeth. The first principal component (PC1) is overlaid on the tooth shape, illustrating its orientation with respect to the horizontal axis. This angle is used to calculate the tooth's rotation angle ($\theta$) for generating oriented bounding boxes (OBB).

Figure 10 illustrates segmentation performance through the visualization of predictions compared to the ground truth, and it also presents the segmentation results using both horizontal bounding boxes and oriented bounding boxes.

So far, the evaluation is based on the entire dataset. We calculate the mean evaluation matrices for each individual tooth label in the entire dataset. However, this approach sometimes does not accurately reflect the performance of individual tooth labels. Therefore, we need to evaluate the 32 individual tooth labels separately. As shown in Figure 11, we generate a radar chart to demonstrate the performance of each label. We observe that tooth labels 4, 8, 22, 24, and 32 show relatively poor performance compared to the other labels in terms of DSC and RIoU. Their DSC scores range from 80% to 85%, and their RIoUs range from 75% to 80%. In contrast, tooth labels 1–2, 9–10, 18–19, 20–21, and 27–29 demonstrate relatively high performance with DSC scores ranging from 90% to 95% and RIoU ranging from 85% to 90%.

Next, we perform categorical analysis. As shown in Figure 2, 32 teeth are divided into four categories: incisors, canines, premolars, and molars. Each of these categories is further divided into two groups: upper and lower. So, in total, we have eight categories. Categorical analysis is presented in Table 4. It is observed that incisors and canines, for both upper and lower jaws, demonstrate good performance. Lower premolars also exhibit good performance. These five categories achieve a DSC of >88% and an RIoU of >83%. However, upper premolars, along with both upper and lower molars, show relatively poor performance, with a DSC of ≤86% and an RIoU of ≤80%.

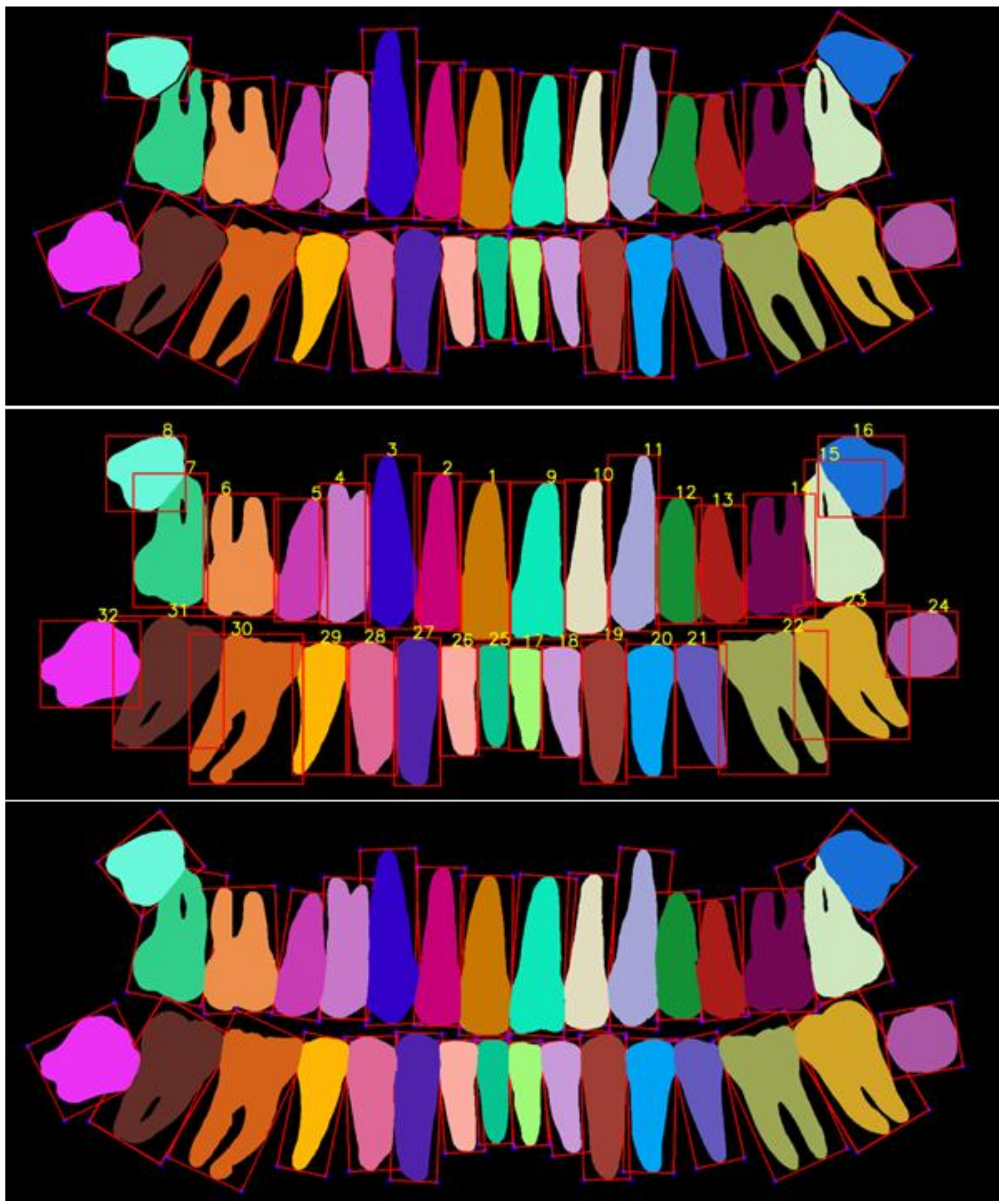

**Figure 10.** Segmentation results: (**top**) Ground truth, (**middle**) Segmentation results with HBB, and (**bottom**) Segmentation results with OBB. Separate colors are chosen for each tooth.

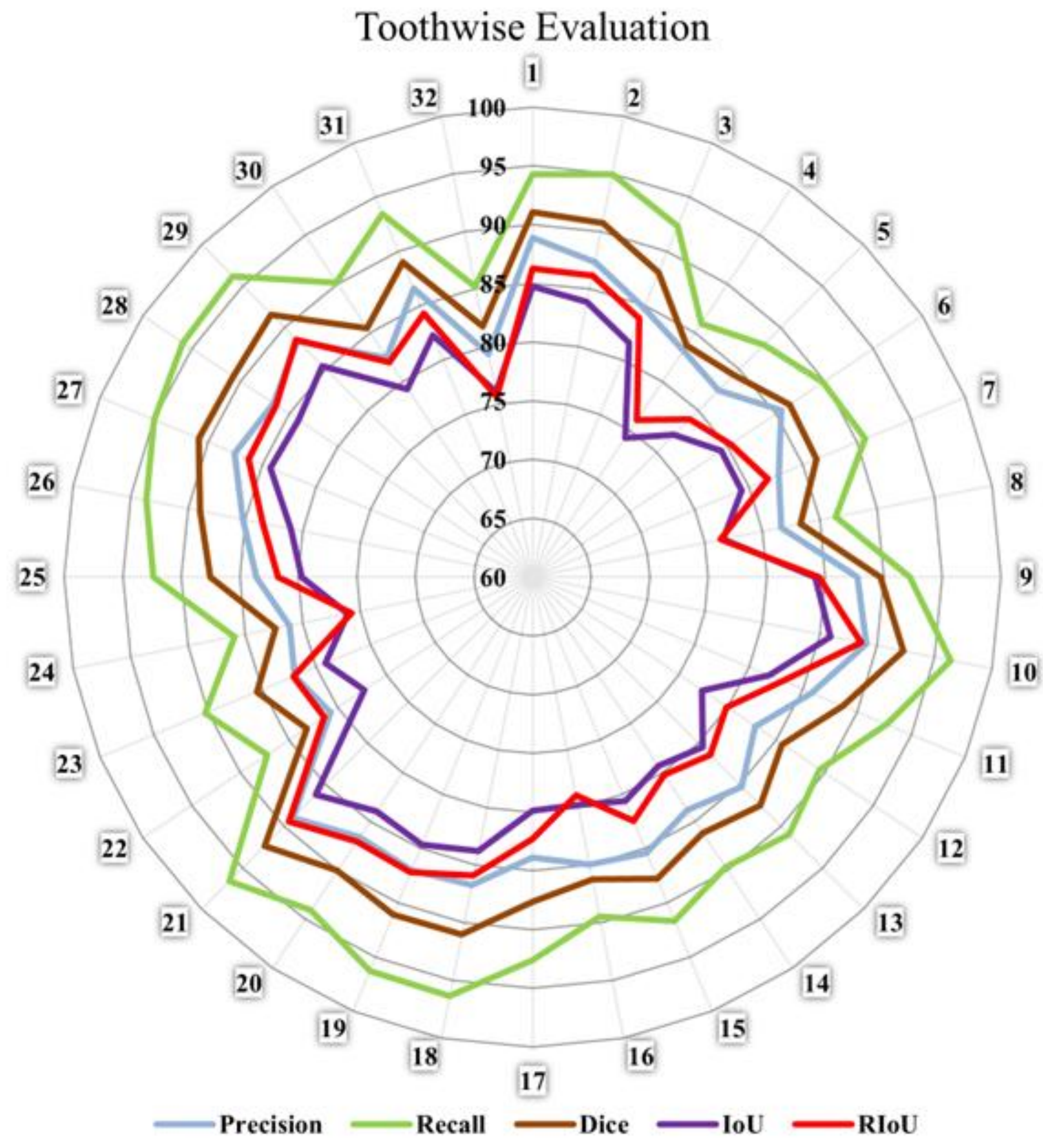

**Figure 11.** Radar diagram to demonstrate tooth-wise evaluation.

**Table 4.** Categorical evaluation of teeth.

| Label | Precision | Recall | DSC | IoU | RIoU |
|---|---|---|---|---|---|
| Upper incisors | 88.25 | 94.45 | 90.95 | 84.66 | 86.39 |
| Lower incisors | 84.84 | 93.76 | 88.77 | 81.11 | 83.35 |
| Upper canine | 85.31 | 92.54 | 88.36 | 81.73 | 83.47 |
| Lower canine | 87.27 | 95.64 | 90.96 | 84.44 | 86.74 |
| Upper premolars | 83.43 | 88.58 | 85.29 | 77.30 | 79.09 |
| Lower premolars | 87.65 | 95.69 | 91.19 | 84.90 | 87.84 |
| Upper molars | 84.04 | 89.67 | 86.05 | 79.19 | 80.08 |
| Lower molars | 82.10 | 88.81 | 84.61 | 78.42 | 80.28 |

We identify several factors that influence the model's performance. Lack of instance overlapping, presence of foreign bodies, fuzzy tooth roots, and poor annotations are some of the factors that affect the performance. As shown in Figure 12a,b, there are overlaps between premolar teeth and between premolar and molar teeth in the panoramic X-ray images; however, there is no instance of overlap in the ground truth. Foreign bodies include dental restorations, broken dental instruments, metal fillings, braces, and other metal objects. Figure 12c shows the presence of bracelets and fillings, while Figure 12d shows the presence of fuzzy tooth roots. As shown in Figure 12e, some images are poorly annotated and contain sharp edges and cones. However, when compared to the original data, our model predicts smoother edges, which is more reasonable. While some researchers have made modifications to improve the dataset's quality [37,38] [25,26], in this work, we have used the original dataset as it is publicly available. To better understand the impact of dataset imperfections, we manually reviewed the test set and identified several recurring issues. Specifically, at least 21 images include foreign objects (e.g., braces, caps, fillings, or screws), at least 19 images (excluding foreign objects) exhibit some form of tooth overlaps that are not reflected in the ground truth, and at least 10 images show ambiguous or fuzzy tooth roots, among which 5 also suffer from low overall image quality. Many of the overlaps are observed in category 2. These factors introduce noise and ambiguity during training and evaluation, potentially affecting the model's ability to learn clear tooth boundaries. While a more rigorous assessment would benefit from clinical validation by dental professionals, such an analysis was beyond the current scope of this study.

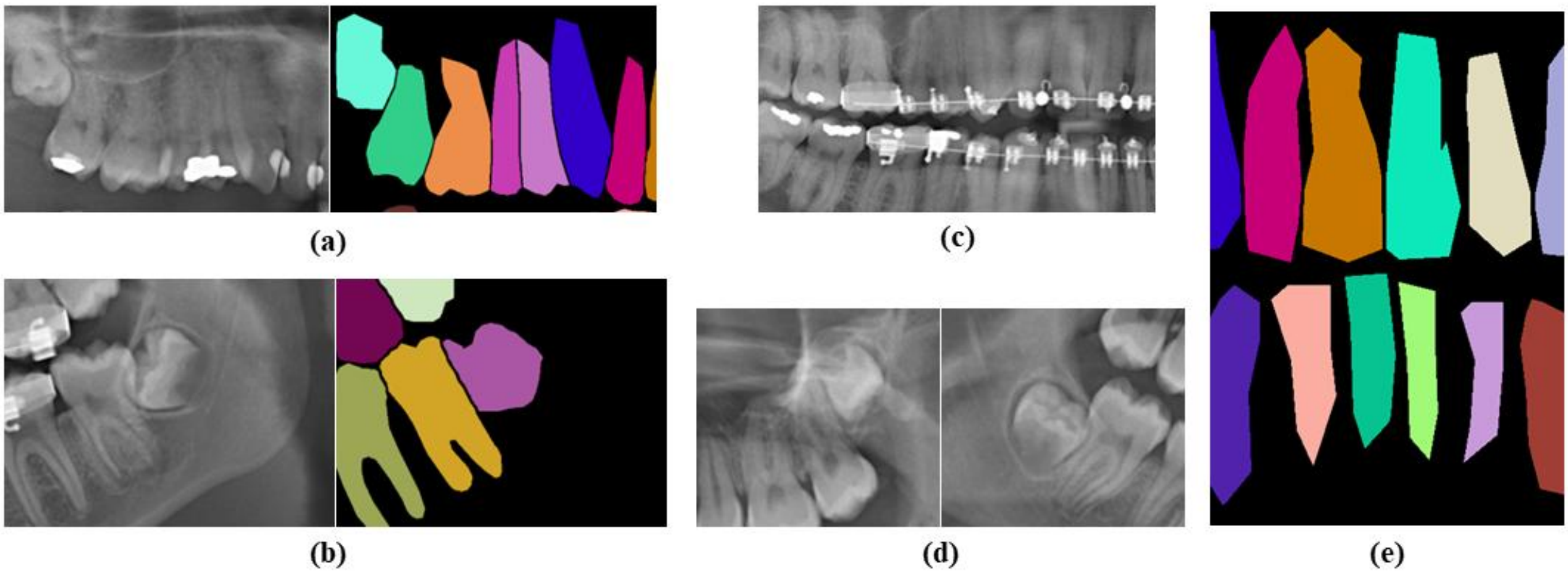

**Figure 12.** Problems in the dataset: (**a**) Overlaps between premolars, and (**b**) overlaps between premolar and molar. However, in the ground truth, there is no instance overlap. (**c**) Presence of braces and fillings, (**d**) fuzzy molar tooth roots, and (**e**) poor annotations with sharp edges and cones.

Furthermore, we calculate the number of false positives and false negatives in OBBs. This is performed to detect missing teeth. There is a total of 3382 teeth in the 111 test images. We train and evaluate our model five times. As shown in Table 3, we do not observe any false positives in any of the runs, and the number of false negatives ranges from three to six. This indicates that the model can be used effectively to detect missing teeth.

The average inference time of the proposed model is approximately 1.2 s per image on Google Colab Pro+, with ~66M model parameters, indicating its suitability for practical clinical use where immediate response is not critical.

## 5. Conclusions

In this paper, we address the critical tasks of teeth segmentation and orientation in the context of oral healthcare. Accurate teeth segmentation holds immense importance for various applications, including orthodontic treatments, clinical diagnoses, surgical procedures, and dental implant planning. We recognize the challenges posed by manual segmentation, especially in lower-resolution images, and the limitations of existing semi-automatic techniques that still require human intervention. Our model architecture includes an EfficientNet-based encoder and a decoder with grid-based attention gates and a parallel spatial and channel squeeze-and-excitation (P-scSE) module. We also introduce oriented bounding boxes (OBB) generation using principal component analysis (PCA). By addressing tooth orientation, a largely unexplored problem, within the same framework as segmentation, our work extends beyond current methodologies, advancing the field toward more comprehensive and clinically useful analysis tools. Our model's accuracy and capabilities open new possibilities for improving dental diagnoses and treatment planning, even when working with clinically standard, moderate-resolution panoramic X-rays. This makes the approach practical for real-world deployment where ultra-high-resolution imaging is not always available. Future works include extending our approach to 3D instance segmentation of teeth, which is crucial for applications like orthognathic surgery planning and dental implant placement, would be a valuable direction. In addition, we plan to explore advanced data augmentation techniques, including geometric transformations, elastic deformations, and intensity perturbations, to improve the robustness and generalization of the model. We also aim to adapt the proposed framework to other dental imaging modalities, such as bitewing and periapical X-rays, with appropriate adjustments for their localized views and anatomical characteristics.

## References

1. Polizzi, A.; Quinzi, V.; Ronsivalle, V.; Venezia, P.; Santonocito, S.; Lo Giudice, A.; Leonardi, R.; Isola, G. Tooth automatic segmentation from CBCT images: A systematic review. *Clin. Oral Investig.* **2023**, *27*, 3363–3378.
2. Luo, D.; Zeng, W.; Chen, J.; Tang, W. Deep learning for automatic image segmentation in stomatology and its clinical application. *Front. Med. Technol.* **2021**, *3*, 767836.
3. Huang, Y.C.; Chen, C.A.; Chen, T.Y.; Chou, H.S.; Lin, W.C.; Li, T.C.; Juan, J.J.; Lin, S.Y.; Li, C.W.; Chen, S.L.; et al. Tooth position determination by automatic cutting and marking of dental panoramic X-ray film in medical image processing. *Appl. Sci.* **2021**, *11*, 11904.
4. Dhar, M.K.; Zhang, T.; Patel, Y.; Gopalakrishnan, S.; Yu, Z. FUSegNet: A Deep Convolutional Neural Network for Foot Ulcer Segmentation. *arXiv* **2023**, arXiv:2305.02961.
5. Silva, B.; Pinheiro, L.; Oliveira, L.; Pithon, M. A study on tooth segmentation and numbering using end-to-end deep neural networks. In Proceedings of the 2020 33rd SIBGRAPI Conference on Graphics, Patterns and Images (SIBGRAPI), Porto de Galinhas, Brazil, 25 November 2020.
6. Koch, T.L.; Perslev, M.; Igel, C.; Brandt, S.S. Accurate segmentation of dental panoramic radiographs with U-NETS. In Proceedings of the 2019 IEEE 16th International Symposium on Biomedical Imaging (ISBI 2019), Venice, Italy, 11 July 2019.
7. Zhao, Y.; Li, P.; Gao, C.; Liu, Y.; Chen, Q.; Yang, F.; Meng, D. TSASNet: Tooth segmentation on dental panoramic x-ray images by two-stage attention segmentation network. *Knowl. Based Syst.* **2020**, *206*, 106338.
8. Chen, Q.; Zhao, Y.; Liu, Y.; Sun, Y.; Yang, C.; Li, P.; Zhang, L.; Gao, C. MSLPNet: Multi-scale Location Perception Network for dental panoramic X-ray image segmentation. *Neural Comput. Appl.* **2021**, 33, 10277–10291.
9. Salih, O.; Duffy, K.J. The local ternary pattern encoder–decoder neural network for dental image segmentation. *IET Image Process.* 2022, 16, 1520–1530. https://doi.org/10.1049/ipr2.12416.
10. Hou, S.; Zhou, T.; Liu, Y.; Dang, P.; Lu, H.; Shi, H. Teeth U-net: A segmentation model of dental panoramic X-ray images for context semantics and contrast enhancement. *Comput. Biol. Med.* **2023**, *152*, 106296.
11. Jader, G.; Fontineli, J.; Ruiz, M.; Abdalla, K.; Pithon, M.; Oliveira, L. Deep instance segmentation of teeth in panoramic X-ray images. In Proceedings of the 2018 31st SIBGRAPI Conference on Graphics, Patterns and Images (SIBGRAPI), Parana, Brazil, 17 January 2018.
12. Rubiu, G.; Bologna, M.; Cellina, M.; Cè, M.; Sala, D.; Pagani, R.; Mattavelli, E.; Fazzini, D.; Ibba, S.; Papa, S.; Ali, M. Teeth segmentation in panoramic dental x-ray using mask regional convolutional Neural Network. *Appl. Sci.* **2023**, *13*, 7947.
13. Panetta, K.; Rajendran, R.; Ramesh, A.; Rao, S.P.; Agaian, S. Tufts Dental Database: A multimodal panoramic X-ray dataset for Benchmarking Diagnostic Systems. *IEEE J. Biomed. Health Inform.* **2022**, *26*, 1650–1659.
14. Helli, S.; Hamamcı, A. Tooth instance segmentation on panoramic dental radiographs using u-nets and morphological processing. *Düzce Üniversitesi Bilim Ve Teknol. Derg.* **2022**, 10, 39–50.
15. El Bsat, A.R.; Shammas, E.; Asmar, D.; Sakr, G.E.; Zeno, K.G.; Macari, A.T.; Ghafari, J.G. Semantic segmentation of maxillary teeth and palatal rugae in two-dimensional images. *Diagnostics* **2022**, *2*, 2176.

16. Wathore, S.; Gorthi, S. Bilateral symmetry-based augmentation method for improved tooth segmentation in panoramic X-rays. *Pattern Recognit. Lett.* **2025**, *188*, 1–7.
17. Brahmi, W.; Jdey, I. Automatic tooth instance segmentation and identification from panoramic X-Ray images using deep CNN. *Multimed. Tools Appl.* **2024**, *83*, 55565–55585.
18. Turk, M.; Pentland, A. Eigenfaces for recognition. *J. Cogn. Neurosci.* **1991**, *3*, 71–86.
19. Tan, M.; Le, Q.V. EfficientNet: Rethinking model scaling for convolutional neural networks. In Proceedings of the 36th International Conference on Machine Learning (ICML), Long Beach, CA, USA, 9–15 June 2019; pp. 10691–10700.
20. Tan, M.; Chen, B.; Pang, R.; Vasudevan, V.; Sandler, M.; Howard, A.; Le, Q.V. Mnasnet: Platform-aware neural architecture search for mobile. In Proceedings of the IEEE/CVF Conference on Computer Vision and Pattern Recognition, Long Beach, CA, USA, 15–20 June 2019.
21. Oktay, O.; Schlemper, J.; Folgoc, L.L.; Lee, M.; Heinrich, M.; Misawa, K.; Mori, K.; McDonagh, S.; Hammerla, N.Y.; Kainz, B.; et al. Attention U-Net: Learning where to look for the pancreas. *arXiv* **2018**, arXiv:1804.03999.
22. Hu, J.; Shen, L.; Sun, G. Squeeze-and-excitation networks. In Proceedings of the 2018 IEEE/CVF Conference on Computer Vision and Pattern Recognition, Salt Lake City, UT, USA, 18–23 June 2018.
23. Roy, A.G.; Navab, N.; Wachinger, C. Recalibrating fully convolutional networks with spatial and channel 'squeeze and excitation' blocks. *IEEE Trans. Med. Imaging* **2019**, 38, 540–549.
24. Dhar, M.K.; Deb, M. S-R2F2U-Net: A single-stage model for teeth segmentation. *Int. J. Biomed. Eng. Technol.* **2024**, *46*, 81–100.
25. Kingma, D.P.; Ba, J. Adam: A method for stochastic optimization. *arXiv* **2014**, arXiv:1412.6980.
26. Lin, T.Y.; Goyal, P.; Girshick, R.; He, K.; Dollár, P. Focal loss for dense object detection. *IEEE Trans. Pattern Anal. Mach. Intell.* **2020**, *42*, 318–327.
27. Xie, E.; Wang, W.; Yu, Z.; Anandkumar, A.; Alvarez, J.M.; Luo, P. SegFormer: Simple and efficient design for semantic segmentation with transformers. *Adv. Neural Inf. Process. Syst.* **2021**, *34*, 12077–12090.
28. Cao, H.; Wang, Y.; Chen, J.; Jiang, D.; Zhang, X.; Tian, Q.; Wang, M. Swin-unet: Unet-like pure transformer for medical image segmentation. In Proceedings of the European Conference on Computer Vision, Tel Aviv, Israel, 23–27 October 2022; Springer Nature: Cham, Switzerland, 2022.
29. Lin, T.; Wang, Y.; Liu, X.; Qiu, X. A survey of Transformers. *AI Open* **2022**, *3*, 111–132.
30. Liu, Y.; Sangineto, E.; Bi, W.; Sebe, N.; Lepri, B.; Nadai, M. Efficient training of visual transformers with small datasets. In Proceedings of the Advances in Neural Information Processing Systems (NeurIPS), Online, 6–14 December 2021; Volume 34, pp. 23818–23830.
31. Chen, L.C.; Zhu, Y.; Papandreou, G.; Schroff, F.; Adam, H. Encoder-decoder with atrous separable convolution for Semantic Image segmentation. In Proceedings of the Computer Vision–ECCV 2018, Munich, Germany, 8–14 September 2018; pp. 833–851.
32. Lin, T.Y.; Dollár, P.; Girshick, R.; He, K.; Hariharan, B.; Belongie, S. Feature Pyramid Networks for Object Detection. In Proceedings of the 2017 IEEE Conference on Computer Vision and Pattern Recognition (CVPR), Honolulu, HI, USA, 21–26 July 2017.
33. Fan, T.; Wang, G.; Li, Y.; Wang, H. Ma-net: A multi-scale attention network for liver and tumor segmentation. *IEEE Access* **2020**, *8*, 179656–179665.
34. Chaurasia, A.; Culurciello, E. LinkNet: Exploiting encoder representations for efficient semantic segmentation. In Proceedings of the 2017 IEEE Visual Communications and Image Processing (VCIP), St. Petersburg, FL, USA, 10–13 December 2017.
35. Zhao, H.; Shi, J.; Qi, X.; Wang, X.; Jia, J. Pyramid Scene Parsing Network. In Proceedings of the2017 IEEE Conference on Computer Vision and Pattern Recognition (CVPR), Honolulu, HI, USA, 21–26 July 2017.
36. Ronneberger, O.; Fischer, P.; Brox, T. U-net: Convolutional networks for biomedical image segmentation. In Proceedings of the Medical Image Computing and Computer-Assisted Intervention–MICCAI 2015: 18th International Conference, Munich, Germany, 5–9 October 2015; pp. 234–241.
37. Kanwal, M.; Ur Rehman, M.M.; Farooq, M.U.; Chae, D.K. Mask-transformer-based networks for teeth segmentation in panoramic radiographs. *Bioengineering* **2023**, 10, 843.
38. Almalki, A.; Latecki, L.J. Self-supervised learning with masked image modeling for teeth numbering, detection of dental restorations, and instance segmentation in dental panoramic radiographs. In Proceedings of the 2023 IEEE/CVF Winter Conference on Applications of Computer Vision (WACV), Waikoloa, HI, USA, 2–7 January 2023.